\documentclass[english,runningheads,a4paper]{llncs}
\usepackage{amssymb}
\usepackage{enumerate}
\usepackage{color}
\usepackage{amsmath}
\usepackage{graphicx}
\usepackage{listings}
\usepackage{lstautogobble}
\usepackage{hhline}
\usepackage{fancyvrb}
\DefineVerbatimEnvironment{Code}{BVerbatim}{baseline=t}

\usepackage{url}

\usepackage[T2A]{fontenc}  
\usepackage[utf8]{inputenc}
\usepackage[english]{babel}

\begin{document}

\title{RuBQ: A Russian Dataset for Question Answering over Wikidata}

\author{Vladislav Korablinov\inst{1}\thanks{Work done as an intern at JetBrains Research.} \and
Pavel Braslavski \inst{2,3}\orcidID{0000-0002-6964-458X}}

\institute{ITMO University, Saint Petersburg, Russia \\
\email{vladislav.korablinov@gmail.com}\\
\and
JetBrains Research, Saint Petersburg, Russia  \and
Ural Federal University, Yekaterinburg, Russia\\
\email{pbras@yandex.ru}}

\maketitle

\begin{abstract}
The paper presents \textbf{RuBQ}, the first Russian knowledge base question answering (KBQA) dataset. The high-quality dataset consists of 1,500 Russian questions of varying complexity, their English machine translations, SPARQL queries to Wikidata, reference answers, as well as a Wikidata sample of triples containing entities with Russian labels. The dataset creation started with a large collection of question-answer pairs from online quizzes. The data underwent automatic filtering, crowd-assisted entity linking, automatic generation of SPARQL queries, and their subsequent in-house verification. 
The freely available dataset will be of interest for a wide community of researchers and practitioners in the areas of Semantic Web, NLP, and IR, especially for those working on multilingual question answering. The proposed dataset generation pipeline proved to be efficient and can be employed in other data annotation projects. 

\keywords{Knowledge base question answering \and Semantic parsing 
\and Evaluation \and Russian language resources}

\textbf{Resource location:} \url{http://doi.org/10.5281/zenodo.3835913}

\textbf{Project page:} \url{https://github.com/vladislavneon/RuBQ}

\end{abstract}

\section{Introduction}

Question answering (QA) addresses the task of returning a precise and concise answer to a natural language question posed by the user. QA received a great deal of attention both in academia and industry. Two main directions within QA are \textit{Open-Domain Question Answering (ODQA)} and \textit{Knowledge Base Question Answering (KBQA)}. ODQA searches for the answer in a large collection of text documents; the process is often divided into two stages: 1)~retrieval of potentially relevant paragraphs and 2)~spotting an answer span within the paragraph (referred to as \textit{machine reading comprehension, MRC}). In contrast, KBQA uses a \textit{knowledge base} as a source of answers. A knowledge base is a large collection of factual knowledge, commonly structured in subject--predicate--object (SPO) triples, 
for example \texttt{(Vladimir\_Nabokov, spouse, V\'{e}ra\_Nabokov)}.

A potential benefit of KBQA is that it uses knowledge in a distilled and structured form that enables reasoning over facts. In addition, knowledge base structure is inherently language-independent~-- entities and predicates are assigned unique identifiers that are tied to specific languages through labels and descriptions, -- which makes KBs more suitable for multilingual QA. 
The task of KBQA can be formulated as a translation from natural language question into a formal KB query (expressed in SPARQL, SQL, or $\lambda$-calculus). 
In many real-life applications, like in \textit{Jeopardy!} winning IBM Watson~\cite{DeepQA} and major search engines, hybrid QA systems are employed -- they rely on both text document collections and structured knowledge bases. 

High-quality annotated data is crucial for measurable progress in question answering. Since the advent of SQuAD~\cite{squad}, a wide variety of datasets for machine reading comprehension have emerged, see a recent survey~\cite{zhang2019machine}. 
We are witnessing a growing interest in multilingual question answering, which leads to the creation of multilingual MRC datasets~\cite{mlqa,artetxe2019cross,tydiqa}. Multilingual KBQA has received a deal of attention in the literature~\cite{hakimov2017amuse,diefenbach2018question}. However, almost all available KBQA datasets are English, Chinese datasets being an exception. Existing multilingual QALD datasets are rather small.

In this paper we present \textbf{RuBQ} (pronounced [`rubik]) -- \textbf{Ru}ssian Knowledge \textbf{B}ase \textbf{Q}uestions, a KBQA dataset that consists of 1,500 Russian questions of varying complexity along with their English machine translations, corresponding SPARQL queries, answers, as well as a subset of Wikidata covering entities with Russian labels. To the best of our knowledge, this is the first Russian KBQA and semantic parsing dataset. 
To construct the dataset, we started with a large collection of trivia Q\&A pairs harvested on the Web. We built a dedicated recall-oriented Wikidata entity linking tool and verified the obtained answers' candidate entities via crowdsourcing. Then, we generated paths between possible question entities and answer entities and carefully verified them.

The freely available dataset is of interest for a wide community of Semantic Web, natural language processing (NLP), and information retrieval (IR) researchers and practitioners, who deal with multilingual question answering. The proposed dataset generation pipeline proved to be efficient and can be employed in other data annotation projects.

\section{Related work}

\begin{table}[tb]
    \centering
        \caption{KBQA datasets. Target knowledge base (\textbf{KB}): Fb -- Freebase, DBp~-- DBpedia, Wd~-- Wikidata 
        (MSParS description does not reveal the details about the KB associated with the dataset). \textbf{CQ} indicates the presence of complex questions in the dataset. Logical form (\textbf{LF}) annotations: $\lambda$~-- lambda calculus, S -- SPARQL queries, t -- SPO triples. Question generation method (\textbf{QM}): M -- manual generation from scratch, SE -- search engine query suggest API, L -- logs, T+PP -- automatic generation of question surrogates based on templates followed by crowdsourced paraphrasing, CS -- crowdsourced manual generation based on formal representations,  QZ~-- quiz collections, FA -- fully automatic generation based on templates.}
    \begin{tabular}{lrrlclll}
        \textbf{Dataset} &  \textbf{Year} & \textbf{\#Q} & \textbf{KB}  & \textbf{CQ} & \textbf{LF} & \textbf{QM} & \textbf{Lang}\\
         \hline
        Free917~\cite{free917} & 2013 & 917 & Fb & + & $\lambda$ & M & en\\
        WebQuestions~\cite{webquestions} & 2013 & 5,810 & Fb &  + & -- & SE & en\\
        SimpleQuestions~\cite{simpleq} & 2015 & 108,442 & Fb & -- & t & CS & en\\
        ComplexQuestions~\cite{bao-etal-2016-constraint} & 2016 & 2,100 & Fb & + & -- & L, SE & en\\
        GraphQuestions~\cite{su2016generating} & 2016 &  5,166 & Fb & + & S & T+PP & en \\
        WebQuestionsSP~\cite{yih2016value} & 2016 & 4,737 & Fb & + & S & SE & en \\
        SimpleQuestions2Wikidata~\cite{simpleQ2wikidata} & 2017 &  21,957 & Wd & -- &  t & CS & en \\
        30M Factoid QA Corpus~\cite{30Mfactoids} & 2017 & 30M & Fb & -- & t & FA & en \\
        LC QuAD~\cite{lc_quad1} & 2017 & 5,000 & DBp & + & S & T+PP & en \\ 
        ComplexWebQuestions~\cite{talmor2018web} & 2018 & 34,689 & Fb & + & S & T+PP & en \\ 
        ComplexSequentialQuestions~\cite{saha2018complex} & 2018 & 1.6M & Wd & + & -- & M+CS+FA & en\\
        QALD9~\cite{QALD9} & 2018 & 558 & DBp & + & S & L & mult\\
        LC-QuAD 2.0~\cite{lc-quad20} & 2019 & 30,000 & DBp, Wd & + & S & T+PP & en\\
        FreebaseQA\cite{freebaseqa} & 2019 & 28,348 & Fb & + & S & QZ & en \\
        MSParS~\cite{duan2019overview} & 2019 & 81,826 & -- & + & $\lambda$ &  T+PP & zh\\
        CFQ~\cite{CompositionGoogle} & 2020 & 239,357 & Fb & + & S & FA & en \\
        RuBQ (this work) & 2020 & 1,500 & Wd & + & S & QZ & ru \\
    \end{tabular}
    \label{tab:KBQA_datasets}
\end{table}

Table~\ref{tab:KBQA_datasets} summarizes the characteristics of KBQA datasets that have been developed to date. These datasets vary in size, underlying knowledge base, presence of questions' logical forms and their formalism, question types and sources, as well as the language of the questions.

The questions of the earliest Free917 dataset~\cite{free917} were generated by two people without consulting a knowledge base, the only requirement was a diversity of questions' topics; each question is provided with its logical form to query Freebase. Berant et al.~\cite{webquestions} created  WebQuestions dataset that is significantly larger but does not contain questions' logical forms. Questions were collected through Google suggest API: authors fed parts of the initial question to the API and repeated the process with the returned questions until 1M questions were reached. After that, 100K randomly sampled questions were presented to MTurk workers, whose task was to find an answer entity in Freebase. Later studies have shown that only two-thirds of the questions in the dataset are completely correct; many questions are ungrammatical and ill-formed~\cite{yih2016value,wu2020perq}. Yih et al.~\cite{yih2016value} enriched 81.5\% of WebQuestions with SPARQL queries and demonstrated that semantic parses substantially improve the quality of KBQA. They also showed that semantic parses can be obtained at an acceptable cost when the task is broken down into smaller steps and facilitated by a handy interface. Annotation was performed by five people familiar with Freebase design, which hints at the fact that the task is still too tough for crowdsourcing. WebQuestions were used in further studies aimed to generate complex questions~\cite{bao-etal-2016-constraint,talmor2018web}. 

SimpleQuestions~\cite{simpleq} is the largest manually created KBQA dataset to date. Instead of providing logical parses for existing questions, the approach explores the opposite direction: based on formal representation, a natural language question is generated by crowd workers. 
First, the authors sampled SPO triples from a Freebase subset, favoring non-frequent subject--predicate pairs. Then, the triples were presented to crowd workers, whose task was to generate a question about the subject, with the object being the answer. This approach doesn't guarantee that the answer is unique~-- Wu et al.~\cite{wu2020perq} estimate that SOTA results on the dataset (about 80\% correct answers) reach its upper bound, since the rest of the questions are ambiguous and cannot be answered precisely. The dataset was used for the fully automatic generation of a large collection of natural language questions from Freebase triples with neural machine translation methods~\cite{30Mfactoids}. Dieffenbach et al.~\cite{simpleQ2wikidata} succeeded in a semi-automatic matching of about one-fifth of the dataset to Wikidata.

The approach behind FreebaseQA dataset~\cite{freebaseqa} is the closest to our study~-- it builds upon a large collection of trivia questions and answers (borrowed largely from TriviaQA dataset for reading comprehension~\cite{triviaqa}). Starting with about 130K Q\&A pairs, the authors run NER over questions and answers, match extracted entities against Freebase, and generate paths between entities. Then, human annotators verify automatically generated paths, which resulted in about 28K items marked relevant. Manual probing reveals that many questions' formal representations in the dataset are not quite precise. For example, the question \texttt{eval-25}: \textit{Who captained the Nautilus in 20,000 Leagues Under The Sea?} is matched with the relation \textit{book.book.characters} that doesn't represent its meaning and leads to multiple answers along with a correct one (\textit{Captain Nemo}). Our approach differs from the above in several aspects. We implement a recall-oriented IR-based entity linking since many questions involve general concepts that cannot be recognized by off-the-shelf NER tools. After that, we verify answer entities via crowdsourcing. Finally, we perform careful in-house verification of automatically generated paths between question and answer entities in KB. We can conclude that our pipeline leads to a more accurate representation of questions' semantics.    

The questions in the KBQA datasets can be \textit{simple}, i.e. corresponding to a single fact in the knowledge base, or \textit{complex}. Complex questions require a combination of multiple facts to answer them. 
WebQuestions consists of 85\% simple questions; SimpleQuestions and 30M factoid QA Corpus contain only simple questions. Many studies~\cite{lc-quad20,bao-etal-2016-constraint,duan2019overview,talmor2018web,saha2018complex,CompositionGoogle} purposefully target complex questions. 

The majority of datasets use Freebase~\cite{freebase} as target knowledge base. Freebase was discontinued and exported to Wikidata~\cite{freebase2wikidata}; the latest available Freebase dump dates back to early 2016. Three collections~\cite{QALD9,lc_quad1,lc-quad20} use DBpedia~\cite{dbpedia}. Newer datsets~\cite{freebase2wikidata,saha2018complex,lc-quad20} use Wikidata~\cite{wikidata}, which is much larger, up-to-date, and has more multilingual labels and descriptions.  
The majority of datasets, where natural language questions are paired with logical forms, employ SPARQL as a more practical and immediate option compared to lambda calculus. 

Existing KBQA datasets are almost exclusively English, with Chinese MSParS dataset being an exception~\cite{duan2019overview}. QALD-9~\cite{QALD9}, the latest edition of QALD shared task,\footnote{See overview of previous QALD datasets in \cite{usbeck2019benchmarking}.} contains questions in 11 languages: English, German, Russian, Hindi, Portuguese, Persian, French, Romanian, Spanish, Dutch, and Italian. The dataset is rather small; at least Russian questions appear to be non-grammatical machine translations. 

There are several studies on knowledge base question generation~\cite{30Mfactoids,elsahar2018zero,indurthi2017generating,CompositionGoogle}. These works vary in the amount and form of supervision, as well as the structure and the complexity of the generated questions. However, automatically generated questions are intended primarily for training; the need for high-quality, human-annotated data for testing still persists. 

\section{Dataset Creation}

Following previous studies~\cite{freebaseqa,triviaqa}, we opted for quiz questions that can be found in abundance online along with the answers. These questions are well-formed and diverse in terms of properties and entities, difficulty, and vocabulary, although we don't control these properties directly during data processing and annotation.  

The dataset generation pipeline consists of the following steps: 1) data gathering and cleaning; 2) entity linking in answers and questions; 3) verification of answer entities by crowd workers; 4) generation of paths between answer entities and question candidate entities; 5) in-house verification/editing of generated paths. In parallel, we created a Wikidata sample containing all entities with Russian labels. This snapshot mitigates the problem of Wikidata's dynamics -- a reference answer may change with time as the knowledge base evolves. In addition, the smaller dataset lowers the threshold for KBQA experiments. In what follows we elaborate on these steps.      

\subsection{Raw Data}

We mined about 150,000 Q\&A pairs from several open Russian quiz collections on the Web.\footnote{\url{http://baza-otvetov.ru}, \url{http://viquiz.ru}, and others.} 
We found out that many items in the collection aren't actual factoid questions, for example, cloze quizzes (\textit{Leonid Zhabotinsky was a champion of Olympic games in \textellipsis [Tokyo]}\footnote{Hereafter English examples are translations from original Russian questions and answers.}), crossword, definition, and multi-choice questions, as well as puzzles (\textit{Q: There are a green one, a blue one, a red one and an east one in the white one. What is this sentence about? A: The White House}). We compiled a list of Russian question words and phrases and automatically removed questions that don't contain any of them. We also removed duplicates and crossword questions mentioning the number of letters in the expected answer. This resulted in 14,435 Q\&A pairs.

\subsection{Entity Linking in Answers and Questions}

We implemented an IR-based approach for generating Wikidata entity candidates mentioned in answers and questions. First, we collected all Wikidata entities with Russian labels and aliases. We filtered out Wikimedia disambiguation pages, dictionary and encyclopedic entries, Wikimedia categories, Wikinews articles, and Wikimedia list articles. We also removed uninformative entities with less than four outgoing relations. These steps resulted in 4,114,595 unique entities with 5,430,657 different labels and aliases. 

After removing punctuation, we indexed the collection with Elasticsearch using built-in tokenization and stemming. Each text string (question or answer) produces three types of queries to the Elasticsearch index: 1)~all token trigrams; 2)~capitalized bigrams (many named entities follow this pattern, e.g. \textit{Alexander Pushkin}, \textit{Black Sea}); and 3)~free text query containing only nouns, adjectives, and numerals from the original string. N-gram queries (types 1 and 2) are run as phrase queries, whereas recall-oriented free text queries (type 3) are executed as Elasticsearch fuzzy search queries. Results of the latter search are re-ranked using a combination of BM25 scores from Elasticsearch and page view statistics of corresponding Wikipedia articles. Finally, we combine search results preserving the type order and retain Top-10 results for further processing. The proposed approach effectively combines precision- (types 1 and 2) and recall-oriented (type~3) processing.

\subsection{Crowdsourcing Annotations}

\begin{figure}[t]
    \center
    \includegraphics[width=\textwidth]{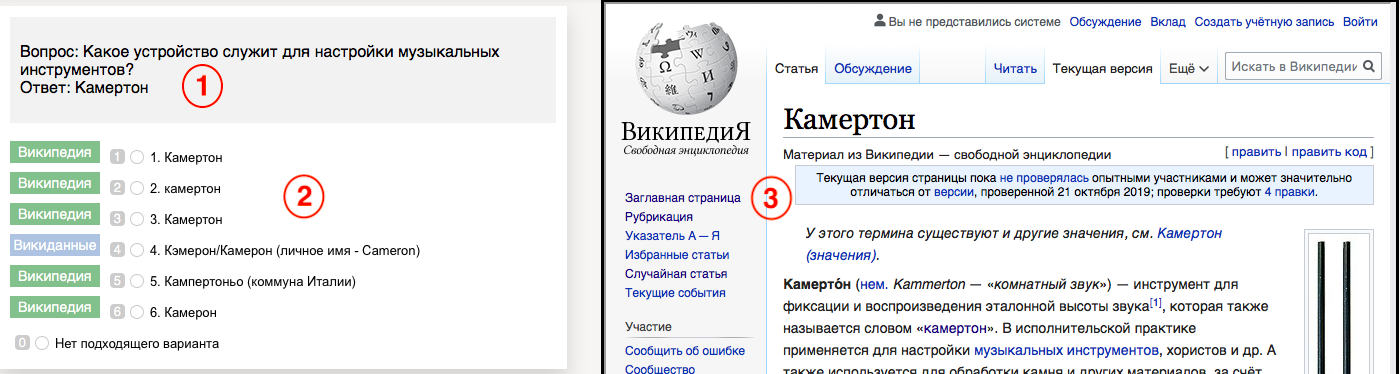}
    \caption{Interface for crowdsourced entity linking in answers: 1 -- question and answer; 2~-- entity candidates; 3 -- Wikpedia page for a selected entity from the list of candidates (in case there is no associated Wikipedia page, the Wikidata item is shown).}
    \label{fig:toloka}
\end{figure}

Entity candidates for answers obtained through the entity linking described above were verified on Yandex.Toloka crowdsourcing platform.\footnote{\url{https://toloka.yandex.com/}} Crowd workers were presented with a Q\&A pair and a ranked list of candidate entities. In addition, they could consult a Wikipedia page corresponding to the Wikidata item, see Figure~\ref{fig:toloka}. The task was to select a single entity from the list or the \textit{None of the above} option. The average number of candidates on the list is 5.43.

Crowd workers were provided with a detailed description of the interface and a variety of examples. To proceed to the main task, crowd workers had to first pass a qualification consisting of 20 tasks covering various cases described in the instruction. We also included 10\% of honeypot tasks for live quality monitoring. These results are in turn used for calculating confidence of the annotations obtained so far as a weighted majority vote (see details in~\cite{ipeirotis2014repeated}). Confidence value governs overlap in annotations: if the confidence is below 0.85, the task is assigned to the next crowd worker. We hired Toloka workers from the best 30\% cohort according to internal rating. As a result, the average confidence for the annotation is 98.58\%; the average overlap is 2.34; average time to complete a task is 19 seconds.

In total, 9,655 out of 14,435 answers were linked to Wikidata entities. Among the matched entities, the average rank of the correct candidate appeared to be 1.5. The combination of automatic candidate generation and subsequent crowdsourced verification proved to be very efficient. A possible downside of the approach is a lower share of literals (dates and numerical values) in the annotated answers. We could match only a fraction of those answers with Wikidata: Wikidata's standard formatted literals may look completely different even if representing the same value. Out of 1,255 date and numerical answers, 683 were linked to a Wikidata entity such as a particular year. For instance, the answer for \textit{In what year was Immanuel Kant born?} matches \textit{Q6926 (year 1724)}, 
whereas the corresponding Wikidata value is \texttt{"1724-04-22"\^{}\^{}xsd:dateTime}. Although the linkage is deemed correct, this barely helps generate a correct path between question and answer entities. 

\subsection{Path Generation and In-house Annotation}

We applied entity linking described above to the 9,655 questions with verified answers and obtained 8.56 candidate entities per question on average. Next, we generated candidate subgraphs spanning question and answer entities, restricting the length between them by two hops. We examined the questions in the sample and found out that longer distances between question and answer entities are very rare.

We investigated the option of filtering out erroneous question entities using crowdsourcing analogous to answer entity verification. A pilot experiment on a small sample of questions showed that this task is much harder -- we got only 64\% correct matches on a test set. Although the average number of generated paths decreased (from 1.9 to 0.9 and from 6.2 to 3.5 for paths of length one and two, respectively), it also led to losing correct paths for 14\% of questions. Thus, we decided to perform an in-house verification of the generated paths. The work was performed by the authors of the paper.

After sending queries to the Wikidata endpoint, we were able to find chains of length one or two for 3,194 questions; the remaining 6,461 questions were left unmatched. We manually inspected 200 random unmatched questions and found out that only 10 of them could possibly be answered with Wikidata, but the required facts are missing in the KB. 

Out of 2,809 1-hop candidates corresponding to 1,799 questions, 866 were annotated as correct. For the rest 2,328 questions, we verified 3,591 2-hop candidates, but only 55 of them were deemed correct. 279 questions were marked as answerable with Wikidata. To increase the share of complex questions in the dataset, we manually constructed SPARQL queries for them.

Finally, we added 300 questions marked as non-answerable over Wikidata, although their answers are present in the knowledge base. These adversarial examples are akin to unanswerable questions in the second edition of SQuAD dataset~\cite{squad20}. The majority of these questions are unanswerable because required predicates are missing in Wikidata, e.g. \textit{How many bells does the tower of Pisa have? (7)}. In some cases, although both question and answer entities are present, the relation between them is missing, e.g. \textit{What circus was founded by Albert Vilgelmovich Salamonsky in 1880? (Moscow Circus on Tsvetnoy Boulevard)}. The presence of such questions makes the task more challenging and realistic.

\section{RuBQ Dataset}

\subsection{Dataset Statistics}

Our dataset has 1,500 unique questions in total. It mentions 2,357 unique entities~-- 1,218 in questions and 1,250 in answers. There are 242 unique relations in the dataset. The average length of the original questions is 7.99 words (median 7); machine-translated English questions are 10.58 words on average (median 10). 131 questions have more than one correct answer. For 1,154 questions the answers are Wikidata entities, and for 46 questions the answers are literals.

Inspired by a taxonomy of query complexity in LC QuAD 2.0~\cite{lc-quad20}, we annotated obtained SPARQL queries in a similar way. 
The query type is defined by the constraints in the SPARQL query, see Table~\ref{tab:question_types}. Note that some queries have multiple type tags. For example, SPARQL query for the question \textit{How many moons does Mars have?} is assigned \textit{1-hop} and \textit{count} types and therefore isn't simple in terms of SimpleQuestions dataset.

Taking into account RuBQ's modest size, we propose to use the dataset primarily for testing rule-based systems, cross-lingual transfer learning models, and models trained on automatically generated examples, similarly to recent MRC datasets~\cite{tydiqa,artetxe2019cross,mlqa}. We split the dataset into development (300) and test (1,200) sets in such a way to keep a similar distribution of query types in both subsets. 

\begin{table}[t]
    \caption{Query types in RuBQ (\#D/T -- number of questions in development and test subsets, respectively).}
    \begin{tabular}{p{0.15\textwidth}|p{0.1\textwidth}|p{0.75\textwidth}}
        Type & \#D/T & Description \\
        \hline
        1-hop & 198/760 & Query corresponds to a single SPO triple \\ 
        multi-hop & 14/55 & Query's constraint is applied to more than one fact \\ 
        multi-constraint & 21/110 & Query contains more than one SPARQL constraint \\
        qualifier-answer & 1/5 & Answer is a value of a qualifier relation, similar to ``fact with qualifiers'' in LC-QuAD 2.0 \\
        qualifier-constraint & 4/22 & Query poses constraints on qualifier relations; a superclass of ``temporal aspect'' in LC-QuAD 2.0 \\
        reverse & 6/29 & Answer's variable is a subject in at least one constraint \\
        count & 1/4 & Query applies \texttt{COUNT} operator to the resulting entities, same as in LC-QuAD 2.0 \\
        ranking & 3/16 & \texttt{ORDER} and \texttt{LIMIT} operators are applied to the entities specified by constraints, same as in LC-QuAD 2.0 \\
        0-hop & 3/12 & Query returns an entity already mentioned in the questions. The corresponding questions usually contain definitions or entity's alternative names \\ 

        exclusion & 4/18 & Query contains \texttt{NOT IN}, which excludes entities mentioned in the question from the answer\\

        no-answer & 60/240 & Question cannot be answered with the knowledge base, although answer entity may be present in the KB \\
    \end{tabular}
    \label{tab:question_types}
\end{table}

\subsection{Dataset Format}

\begin{table}[t]
    \caption{Examples from the RuBQ dataset. Answer entities' labels are not present in the dataset and are cited here for convenience. Note that the original Q\&A pair corresponding to the third example below contains only one answer -- \textit{geodesist}.}
    \begin{tabular}{p{0.2\textwidth}|p{0.8\textwidth}}
        Question & Who wrote the novel ``Uncle Tom's Cabin''? \\
        \hline
        SPARQL query  & 
        \begin{Code} 
 SELECT ?answer 
 WHERE {
   wd:Q2222 wdt:P50 ?answer
 }
        \end{Code}
        \\
        \hline
        Answers IDs & Q102513 (Harriet Beecher Stowe) \\
        \hline
        Tags & 1-hop \\
        \hhline{==}
        Question & Who played Prince Andrei Bolkonsky in Sergei Bondarchuk's film ``War and Peace''? \\
        \hline
        SPARQL query & 
        \begin{Code}
 SELECT ?answer
 WHERE {
   wd:Q845176 p:P161 
   [ ps:P161 ?answer; pq:P453 wd:Q2737140 ]
 }       
 \end{Code}
 \\
        \hline
        Answers IDs & Q312483 (Vyacheslav Tikhonov) \\
        \hline
        Tags & qualifier-constraint \\
        \hhline{==}
        Question & Who uses a theodolite for work? \\
        \hline
        SPARQL query &
        \begin{Code}
 SELECT ?answer 
 WHERE {
   wd:Q181517 wdt:P366 [ wdt:P3095 ?answer ]
 }   
 \end{Code}
 \\
        \hline
        Answers IDs &
        Q1734662 (cartographer), 
        Q11699606 (geodesist), 
        Q294126 (land surveyor)\\
        \hline
        Tags & multi-hop \\
        % \hline
        % English MT & Someone at work uses a theodolite? \\
    \end{tabular}
    \label{tab:RuBQ_example}
\end{table}

For each entry in the dataset, we provide: the
original question in Russian,
machine-translated English question obtained through Yandex.Translate,\footnote{\url{https://translate.yandex.com/}}  original answer text (may differ textually from the answer entity's label retrieved from Wikidata),
SPARQL query representing the meaning of the question, 
a list of entities in the query, 
a list of relations in the query,
a list of answers (a result of querying the Wikidata subset, see below), 
and a list of query type tags, see Table~\ref{tab:RuBQ_example} for examples.  RuBQ is distributed under CC BY-SA license and is available in JSON format. 

The dataset is accompanied by \texttt{RuWikidata8M} -- a Wikidata sample containing all the entities with Russian labels.\footnote{\url{https://zenodo.org/record/3751761},  project's page on github points here.} It consists of about 212M triples with 8.1M unique entities. As mentioned before, the sample guarantees the correctness of the queries and answers and makes the experiments with the dataset much simpler. For each entity, we executed a series of \texttt{CONSTRUCT} SPARQL queries to retrieve all the truthy statements and all the full statements with their linked data.\footnote{Details about Wikidata statement types can be found here: \url{https://www.mediawiki.org/wiki/Wikibase/Indexing/RDF_Dump_Format#Statement_types}}
We also added all the triples with \texttt{subclass of (P279)} predicate to the sample. This class hierarchy can be helpful for question answering task in the absence of an explicit ontology in Wikidata.
The sample contains Russian and English labels and aliases for all its entities. 

\subsection{Baselines}

\begin{table}[th]
    \caption{DeepPavlov's and WDAqua's top-1 results on RuBQ's answerable and unanswerable questions in the test set, and the breakdown of correct answers by query type.}
    \centering
    \begin{tabular}{l|r|r}
    & DeepPavlov & WDAqua \\ 
    \hline
        \multicolumn{3}{c}{Answerable (960) }\\
    \hline
        \textbf{correct} &  129 & 153  \\
        1-hop & 123 & 136 \\
        1-hop + reverse & 0 & 3  \\
        1-hop + count & 0 & 2 \\
        1-hop + exclusion & 0 & 2 \\
        multi-constraint & 4 & 9 \\
        multi-hop & 1 & 0 \\
        qualifier-constraint & 1 & 0 \\
        qualifier-answer & 0 & 1 \\
        \textbf{incorrect/empty} & 831 & 807 \\
    \hline
        \multicolumn{3}{c}{Unanswerable (240) }\\
    \hline
        \textbf{incorrect} & 65 & 138 \\
        \textbf{empty/not found} & 175 & 102 \\
    \end{tabular}
    \label{tab:deeppavlov}
\end{table}

We provide two RuBQ baselines from third-party systems -- DeepPavlov and WDAqua -- that illustrate two possible approaches to cross-lingual KBQA. 

To the best of our knowledge, the KBQA library\footnote{\url{http://docs.deeppavlov.ai/en/master/features/models/kbqa.html}} from an open NLP framework DeepPavlov~\cite{deeppavlov} is the only freely available KBQA implementation for Russian language.  
The library uses Wikidata as a knowledge base and implements the standard question processing steps: NER, entity linking, and relation detection. According to the developers of the library, they used machine-translated SimpleQuestions and a dataset for zero-shot relation extraction~\cite{levy-etal-2017-zero} to train the model. The library returns a single string or \textit{not found} as an answer. We obtained an answer entity ID using reverse ID-label mapping embedded in the model. If no ID is found, we treated the answer as a literal.

WDAqua~\cite{WDAqua} is a rule-based KBQA system that answers questions in several languages using Wikidata. WDAqua returns a (possibly empty) ranked list of Wikidata item IDs along with corresponding SPARQL queries. We obtain WDAqua's answers by sending RuBQ questions machine-translated into English to its API.\footnote{\url{www.wdaqua.eu/qa}}

WDAqua outperforms DeepPavlov in terms of precision@1 on the answerable subset (16\% vs. 13\%), but demonstrates a lower accuracy on unanswerable questions (43\% vs. 73\%). Table~\ref{tab:deeppavlov} presents detailed results.
In contrast to DeepPavlov, WDAqua returns a ranked list of entities as a response to the query, and for 23 out of 131 questions with multiple correct answers, it managed to perfectly match the set of answers. For eight questions with multiple answers, WDAqua's top-ranked answers were correct, but the lower-ranked ones contained errors. To facilitate different evaluation scenarios, we provide an evaluation script that calculates precision@1, exact match, and precision/recall/F1 measures, as well as the breakdown of results by query types.

\section{Conclusion and Future Work}

We presented RuBQ -- the first Russian dataset for Question Answering over Wikidata. The dataset consists of 1,500 questions, their machine translations into English, and annotated SPARQL  queries. 300 RuBQ questions are unanswerable, which poses a new challenge for KBQA systems and makes the task more realistic. The dataset is based on a collection of quiz questions. The data generation pipeline combines automatic processing, crowdsourced and in-house verification, and proved to be very efficient. The dataset is accompanied by a Wikidata sample of 212M triples that contain 8.1M entities with Russian and English labels, and an evaluation script. The provided baselines demonstrate the feasibility of the cross-lingual approach in KBQA, but at the same time indicate there is ample room for improvements. The dataset is of interest for a wide community of researchers in the fields of Semantic Web, Question Answering, and Semantic Parsing.

In the future, we plan to explore other data sources and approaches for RuBQ expansion: search query suggest APIs as for WebQuestions~\cite{webquestions}, a large question log~\cite{volske2015users}, and Wikidata SPARQL query logs.\footnote{\url{https://iccl.inf.tu-dresden.de/web/Wikidata_SPARQL_Logs/en}} 
We will also address complex questions and questions with literals as answers, as well as the creation of a stronger baseline for RuBQ.

\subsubsection{Acknowledgments.} We thank Mikhail Galkin, Svitlana Vakulenko, Vladimir Kovalenko, Yaroslav Golubev, and Rishiraj Saha Roy for their valuable comments and fruitful discussion on the paper draft. We also thank Pavel Bakhvalov, who helped collect \texttt{RuWikidata8M} sample and contributed to the first version of the entity linking tool. We are grateful to Yandex.Toloka for their data annotation grant. 

\bibliographystyle{splncs04}
\bibliography{bib}

\end{document}